\documentclass{article}

\PassOptionsToPackage{numbers, compress}{natbib}

    \usepackage[final]{neurips_2022}

\usepackage[utf8]{inputenc} %
\usepackage[T1]{fontenc}    %
\usepackage{hyperref}       %
\usepackage{url}            %
\usepackage{booktabs}       %
\usepackage{amsfonts}       %
\usepackage{nicefrac}       %
\usepackage{microtype}      %
\usepackage{xcolor}         %

\usepackage{graphicx}
\usepackage{subfig}
\usepackage{algpseudocode} 
\usepackage{mwe}
\usepackage{array}
\usepackage{amsmath}
\usepackage{amsthm}
\usepackage{enumitem}
\usepackage{cleveref}

\graphicspath{ {./figures/} }

\title{A General-Purpose Neural Architecture\\ for Geospatial Systems}

\author{Nasim Rahaman$^{*, 1, 2}$ $\;$ Martin Weiss$^{*, 1, 6}$ $\;$ Frederik Tr\"auble$^{2}$ $\;$ Francesco Locatello $^{3}$ $\;$ \\ \textbf{Alexandre Lacoste}$^{4}$ $\;$ \textbf{Yoshua Bengio}$^{1, 5, 7}$ $\;$ \\ \textbf{Chris Pal}$^{\ddagger, 1, 6, 7}$ $\;$ \textbf{Li Erran Li} $^{\ddagger, 3}$ $\;$ \textbf{Bernhard Sch\"olkopf} $^{\,\ddagger, 2}$ \\ \\ $^{*, \ddagger}$ Equal contribution, random order. \\
\\ $^{1}$Mila, Quebec AI Institute $\;$ $^{2}$ Max Planck Institute for Intelligent Systems, T\"ubingen \\ $^{3}$ AWS AI $\;$ $^{4}$ ServiceNow Research $\;$ $^{5}$ Universit\'e de Montr\'eal 
$\;$ $^{6}$ Polytechnique Montr\'eal \\ $^{7}$ Canada CIFAR AI Chair
}

\begin{document}

\maketitle

\begin{abstract}
    Geospatial Information Systems are used by researchers and Humanitarian Assistance and Disaster Response (HADR) practitioners to support a wide variety of important applications. However, collaboration between these actors is difficult due to the heterogeneous nature of geospatial data modalities (e.g., multi-spectral images of various resolutions, timeseries, weather data) and diversity of tasks (e.g., regression of human activity indicators or detecting forest fires). In this work, we present a roadmap towards the construction of a general-purpose neural architecture (GPNA) with a geospatial inductive bias, pre-trained on large amounts of unlabelled earth observation data in a self-supervised manner. We envision how such a model may facilitate cooperation between members of the community. 
    We show preliminary results on the first step of the roadmap, where we instantiate an architecture that can process a wide variety of geospatial data modalities and demonstrate that it can achieve competitive performance with domain-specific architectures on tasks relating to the U.N.'s Sustainable Development Goals. 
\end{abstract}

\section{Introduction}

In August 2005, Hurricane Katrina made landfall in New Orleans, causing an estimated \$125 billion dollars of damage and 1,833 fatalities. The response to this disaster required coordination across a large number of actors, which ultimately minimized the effectiveness and timeliness of the overall response. Fast forward to 2022, the field of deep learning has progressed by leaps and bounds. However, traditional deep learning methods can be difficult to train, are specialized to certain tasks, and require large amounts of labelled data, capital, and time investments to be effective. As a consequence, these methods have seen slower adoption by the HADR community: because in a crisis, time is of the essence.

At the same time, large neural models trained with self-supervision on web-scale text and image datasets have made advances in a wide variety of applications, ranging from novel image synthesis to image-and-text dialogue.
However, similar advances have yet to permeate the domain of remote sensing, despite the availability of large amounts of (unlabeled) data and numerous impactful HADR and climate science applications. 
Partially to blame is the fact that remote sensing data spans many modalities like synthetic aperture radar (SAR), multi- and hyper-spectral optical imagery, and LiDAR. 
Further, labelling such data can be difficult and require substantial domain expertise. 
Consequently, solving remote sensing tasks with deep learning often requires neural architectures tailored to a specific sensor modality, making it difficult to share knowledge across tasks. 

However, the recent development of so-called \textit{general-purpose models} presents exciting new opportunities. General-purpose neural architectures like Perceiver IO \citep{perceiverio} are capable of operating on a wide variety of different data modalities, ranging from images and videos to point clouds, audio and natural language. In doing so, such models forego hard-coded inductive biases associated with a specific data modality -- for instance, convolutions for images, recurrence for text and audio, 3D structure for point clouds, and so on. But this is compensated for by their ability to absorb large amounts of training data and benefit from large-scale pre-training, made feasible by advances in compute hardware and software infrastructure. Moreover, such models are in a unique position to exploit patterns spanning multiple modalities (e.g. frames of a video and the corresponding audio), which special-purpose models might miss.

In this work, we identify the synergy between general purpose models and the remote sensing data domain. We envision and lay out a roadmap towards a model that embraces the heterogeneity of data found in the remote sensing domain, and is enabled by the ever increasing volume of unlabelled data produced by an ever increasing number of space-based and aerial remote-sensing platforms. 
Such a model would support diverse tasks, enable collaboration between actors with different data modalities, and allow interaction via natural language in order to specify new tasks and data modalities.  
Finally, we present a concrete implementation of a single architecture that is competitive on a suite of diverse tasks that overlaps with Sustainbench \citep{yeh2021sustainbench}.

\if{1}
\begin{figure}[!htb]
    \centering
    \includegraphics[width=\textwidth]{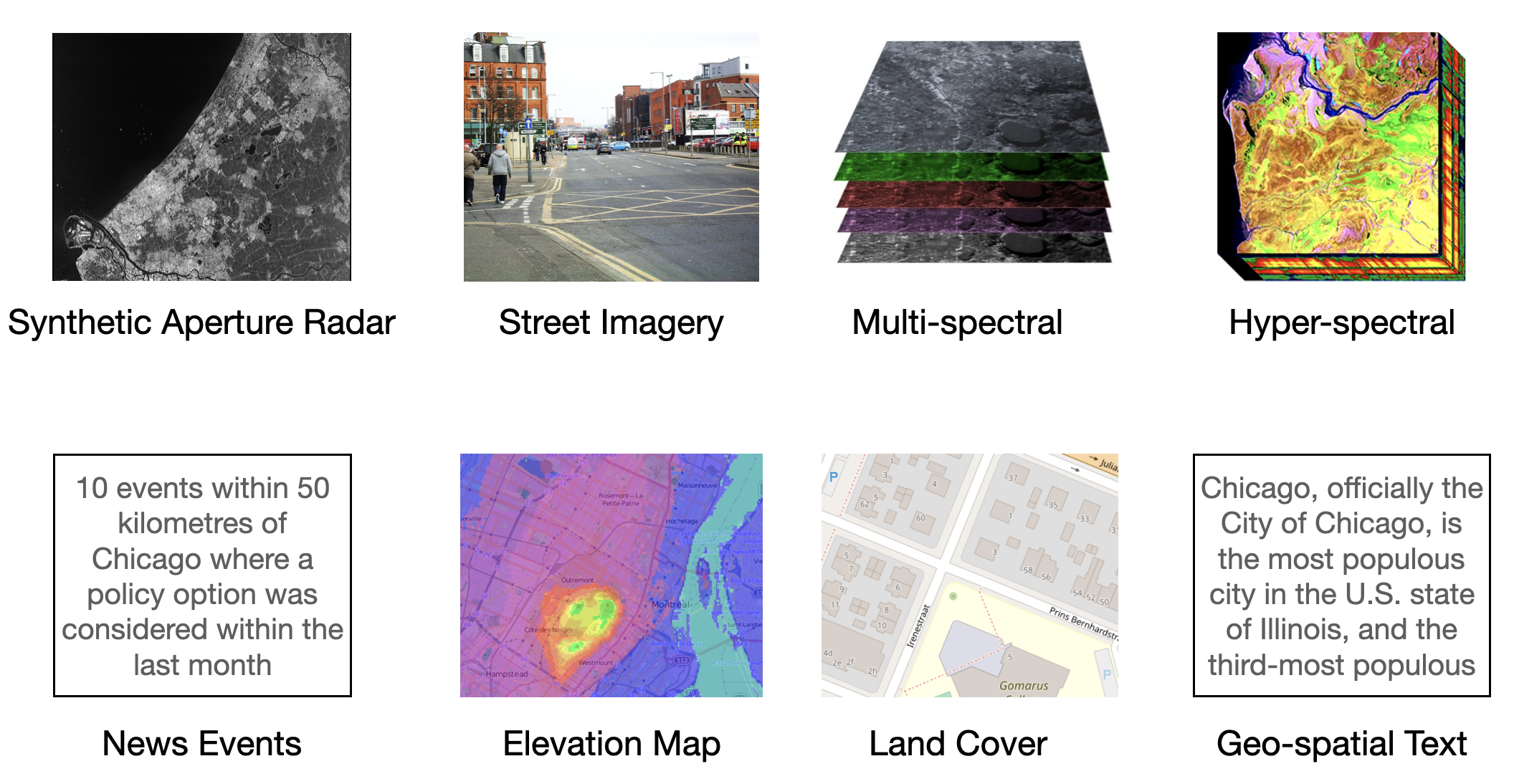}
    \caption{\textbf{Heterogenous Geospatial Data Modalities}. Geospatial data is very diverse, and includes active sensors (SAR), passive sensors (natural images, multispectral, and hyperspectral), derived data (elevation maps), and human annotated information (land cover labels, geo-spatial wikipedia, news events).}
    \label{fig:gamma-data-modalities}
\end{figure}
\fi

\textbf{Our contributions are as follows}. \textbf{(a)} We present a roadmap towards a general-purpose neural model with a geospatial inductive bias. \textbf{(b)} We implement the first step of said roadmap and present promising preliminary results on a suite of sustainability tasks. \textbf{(c)} We outline some of the challenges that we anticipate and discuss possible solutions to address these.

\section{Application context}
The needs of the Humanitarian Assistance and Disaster Response community are diverse in terms of application but are almost universally time-sensitive and safety-critical \citep{lang2020earth, quinn2018humanitarian, ferreira2020monitoring, persello2022deep}. The machine learning community has diligently created tools that can be applied to the wide variety of applications\citep{munawar2022remote, prapas2021deep, shakeel2021detecting}. But these tools are highly-specialized, like a surgeon's scalpel, and they often can only be wielded by certain people intimately familiar with this tool (usually their creators) and only under optimal circumstances. To follow the analogy, the neural architecture we propose is like a swiss army knife for the humanitarian assistance and disaster response community.

\section{A Roadmap Towards General-purpose Geospatial AI Models}
As of today, the \textbf{status quo} of leveraging machine learning methods for remote sensing application is to build and train individual and specialized architectures for each different problem and task domain \citep{yeh2021sustainbench, siamesefcndamage, segnet}. Clearly, this implies a substantial amount of untapped potential of AI given recent progress discussed above. 

We envision a general-purpose neural model that understands both the earth's natural processes and human activities, and can leverage arbitrary sensor data to quickly generalize to new tasks. Such a system would be able to integrate satellite, aerial, and ground-based information from diverse actors and be used by researchers and practitioners to answer important questions for humanitarian and environmental organizations. %
In order for us to obtain unified AI models that can perform the entire spectrum of meaningful remote sensing tasks in a zero-shot manner (i.e. without additional task-specific training), we provide a roadmap with 5 milestone steps to guide further research. 

\begin{figure}[]
    \centering
    \includegraphics[width=1\linewidth]{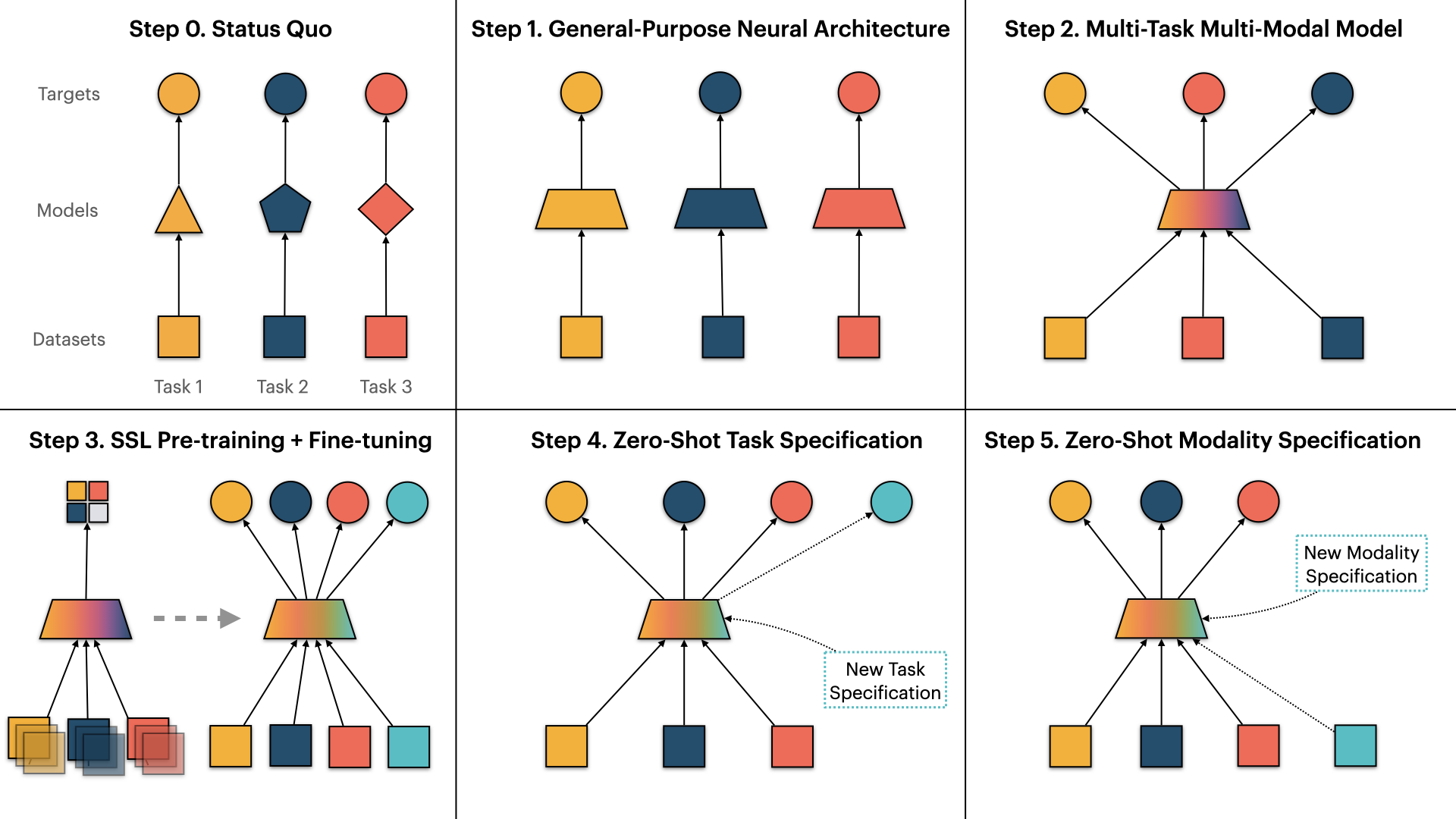}
    \caption{\textbf{Roadmap towards a general-purpose geospatial model}.}
    \label{fig:roadmap}
\vspace{-4mm}
\end{figure}

\begin{enumerate}[topsep=0pt,itemsep=-1ex,partopsep=1ex,parsep=1ex]
\item[\textbf{Step 1}.] Demonstrate that a \textbf{General-Purpose Architecture} can achieve competitive performance across a broad range of geospatial tasks while operating on different geo-spatial and temporal input modalities at the same time. \textbf{The model evaluated in this work overcomes this step}. In a time of crisis, we envision this to already be helpful by expediting the AI workflow in HADR, because the user no longer requires the time and expertise to explore and select an appropriate model for the task at hand. However, the resulting model is still bound to a task, and it might have limited reusability. 

\item[\textbf{Step 2}.] Train a \textbf{multi-task multi-modal model} with a single set of parameters to be competitive across a broad range of geospatial tasks. Overcoming this step should facilitate positive task transfer: for example, learning to segment plant species could improve wildfire behavior prediction, but entails resolving technical and ethical dilemma in early-stopping the model training and thereby trading off between task performances. Nevertheless, this step facilitates collaboration between domain experts who might be interested in different domain-specific tasks, in that they now contribute to training a single, generalist model that profits from the contribution of their peers. However, while this may improve the sample efficiency for the new task, we still expect to require time for data collection and model training.

\item[\textbf{Step 3}.] \textbf{Pre-train} a general-purpose multi-task model on vast amounts unlabeled remote sensing data via self-supervised learning. The pre-trained model should then be able to achieve comparable performance to a newly initialized model on the tasks in steps 1 and 2 with fewer samples and even solve tasks where there is only a tiny amount of labelled data. Like in other domains \citep{t5, GPT3, clip}, this will enable the remote sensing community to perform more tasks faster and with less expensive labelling. Further, it might also enable interactive labelling, where the user reacts to the training procedure by providing labels where the model is not adequately performant \citep{berg2019}. 

\item[\textbf{Step 4}.] Construct a training dataset and enhance the model architecture to enable \textbf{zero-shot generalization to new tasks} with ``language prompts''. The pre-trained model would then not only be able to achieve a competitive zero-shot performance on existing single-task benchmarks, but also additionally leverage any high-level non-visual geo-located information that has been present in text datasets (e.g. the internet).
Concretely, such a model could respond to a long tail of new queries not been seen during training, and as a result the team responding to the crisis would not be required to collect a dataset and train the model to answer this new query. 
Evidence that this type of zero-shot generalization can be achieved has been demonstrated in other settings \citep{promptprogram, flamingo}.

\item[\textbf{Step 5}.] Demonstrate \textbf{zero-shot generalization to new modalities} with ``tokenizer prompts''. 
This is useful for the system to integrate new data modalities as a crisis evolves. 
For example, in the midst of a hurricane it may become necessary to determine which houses are at imminent risk of flooding. If one organization provides an up-to-date heatmap of power outages, then such a model could integrate this new type of data with existing modalities such as altitude, wind-speed, and tidal cycle to improve predictive performance. As a result, collaboration between disaster response organizations and integration of new data becomes much easier.
\end{enumerate}

\section{Method: A General Purpose Neural Architecture for Geospatial Systems}
\label{sec:method}
\begin{figure}[]
    \centering
    \includegraphics[width=\textwidth]{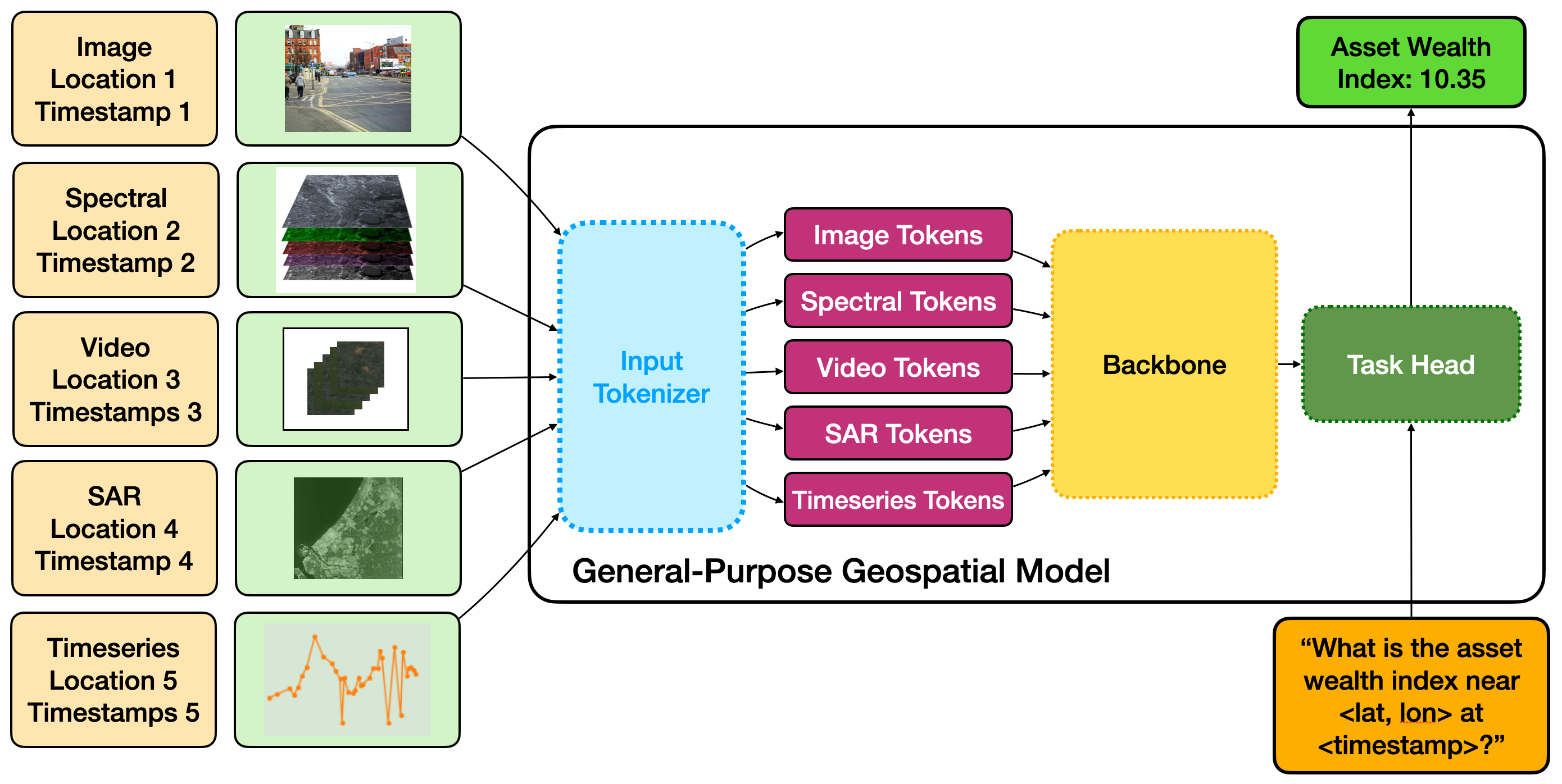}
    \caption{Model Architectural Overview. The General-Purpose Neural Architecture for Geospatial Systems is composed of three primary components. First, the \textbf{input tokenizer} which transforms a variety of input keys (modality type, spatial and temporal information) and values (the captured data) into a common token encoding. Second, the \textbf{model backbone} processes and fuses these tokens. Finally, an input query is provided to the \textbf{task head} with the backbone's output representation to make a prediction.}
    \label{fig:gamma-architecture}
\vspace{-5mm}
\end{figure}

\paragraph{Architectural Overview.} We propose a General-Purpose Neural Architecture (see Figure \ref{fig:gamma-architecture}) that builds on Perceiver IO \citep{perceiverio} and comprises three key components: one or more tokenizers, a backbone and one or more task heads. We provide additional details in the appendix.

\textbf{Tokenizers.} The purpose of a tokenizer is to homogenize the input to the model by converting it to a set of vectors, called tokens, which are then fed to the backbone via a cross-attention mechanism. There are several challenges that arise here. For example, how should we encode multi-spectral satellite imagery with its acquisition location, time, and information about its spectral bands? Or perhaps time-series inputs and SAR (synthetic aperture radar) imagery, including the polarization modes? Addressing these issues required us to innovate in the design of positional, temporal, and spectral encodings, as well as the use of convolutional preprocessors, details of which we provide in the appendix. The proposed GPNA supports the simultaneous use of multiple tokenizers to encode multi-modal data samples. In Step 5, we proposed the development of a general-purpose \textit{tokenizer} which could be conditioned on natural language to encode new data modalities.

\textbf{Backbone.} The backbone is the main computational trunk of the model. It is based on the Perceiver IO \citep{perceiverio}, a recently introduced GPNA that operates as following. It ingests the set of vector-valued tokens produced by the tokenizer and produces another set of vector-valued \textit{latent representations} (where the two sets may have different cardinalities). The input tokens are fed to a cross-attention mechanism, where they are used as keys and values, whereas the queries of this cross attention mechanism are freely learnable parameters. The output of the initial cross attention layer is ingested by a stack of self attention layers, which produces the output latent representations. Crucially, the backbone is shared between all data modalities and tasks (see below).

\textbf{Task Heads.} A task head consumes the latent representation vectors produced by the backbone via another cross-attention mechanism followed by a MLP to ultimately produce the model output. The proposed architecture supports the use of multiple task-heads simultaneously, all consuming the same backbone output features.  We have implemented a classification, regression and semantic segmentation head, and in our experiments we sometimes instantiate multiple heads of the same type (e.g., we train a single model with two segmentation heads to perform the Field Delineation and Farm Area tasks).  Several more task heads are feasible, including ones for object detection, spatial density prediction and for responding to natural language queries \citep{flamingo}.

\section{Experiments}

\paragraph{Experimental setup.}
We argue that \textbf{our model achieves Step 1 of the roadmap}. We evaluate our method against a variety of tasks from SustainBench \citep{yeh2021sustainbench}, a benchmark that requires the prediction of targets associated with the UN Sustainable Development Goals (including poverty, hunger, health, well-being, quality education, clean water, and sanitation) from a wide variety of inputs that are typical for remote sensing datasets such as multi-spectral images, time-series, and SAR data. The top of the SustainBench leaderboard is diverse, including KNNs, ResNets \citep{resnet}, and U-Nets\citep{unet}. Across all tasks we use the same general-purpose neural architecture described in Section \ref{sec:method}. As baselines, we compare our models' task performance with special-purpose models as reported in \cite{yeh2021sustainbench}.

\paragraph{Results.}
We present a complete account of task performance in Table \ref{tab:results}. Overall, our proposed GPNA matches or outperforms special-purpose model performances reported in \cite{yeh2021sustainbench} on 8 of 12 tasks. We train a GPNA model \textbf{from scratch} on each task individually and compare against similarly trained models on the SustainBench leaderboard. In some cases, we see that our architecture dramatically outperforms existing results such as in the Temporal Poverty prediction task, where the model must predict the change in asset wealth index over time given two multi-spectral images. In this case, part of the improvement is due to this particular SustainBench task providing 5 times more samples than those used to obtain the results reported on their leaderboard. However, part of the improvement could also be due to our model using the temporal information to gauge the magnitude of the change. Similarly, our model outperforms on the crop yield task, which we hypothesize is due to its ability to better incorporate important spatiotemporal information than alternative methods. The proposed model currently performs poorly on segmentation tasks. We believe the spatiotemporal encodings used to query the output representation do not sufficiently encode the relative pixel positions, but that with additional regularization and longer training these can become competitive with U-Nets.

\begin{table}[htb]
\resizebox{\textwidth}{!}{
\begin{tabular}{@{}lcccc@{}}
\toprule
                   & \textbf{\begin{tabular}[c]{@{}l@{}}GPNA\\ Single-task\end{tabular}} & \textbf{\begin{tabular}[c]{@{}l@{}}Status Quo\\ Performance\end{tabular}} & \textbf{\begin{tabular}[c]{@{}l@{}}Status Quo Method\end{tabular}} & \textbf{Metric} \\ \midrule
Poverty (Spatial)  & \textbf{0.63}                                                       & \textbf{0.63}                                                                  & KNN                                                                       & R² (↑)          \\
Poverty (Temporal) & \textbf{0.88}                                                       & 0.35                                                                           & ResNet-18                                                                 & R² (↑)          \\
Land Cover         & \textbf{0.60}                                                       & 0.58                                                                           & \begin{tabular}[c]{@{}l@{}}Tile2Vec  ResNet-50\end{tabular}             & Accuracy (↑)    \\
Has Brick Kiln     & \textbf{0.96}                                                       & 0.94                                                                           & ResNet-50                                                                 & Accuracy (↑)    \\
Crop Yield         & \textbf{0.09}                                                       & 0.37                                                                           & CNN + GP                                                                  & RMSE (↓)        \\
Field Delineation  & 0.33                                                                & \textbf{0.56}                                                                  & Spatial U-Net                                                             & DICE  (↑)       \\
Farm Area          & 0.67                                                                & \textbf{0.72}                                                                  & Spatial U-Net                                                             & F1 (↑)          \\
Women's Education  & \textbf{0.54}                                                       & 0.26                                                                           & KNN                                                                       & R²  (↑)         \\
Women's BMI        & 0.32                                                                & \textbf{0.42}                                                                  & KNN                                                                       & R² (↑)          \\
Under 5 Mortality  & \textbf{0.02}                                                       & 0.01                                                                           & KNN                                                                       & R² (↑)          \\
Sanitation Index   & \textbf{0.472}                                                      & 0.36                                                                           & KNN                                                                       & R² (↑)          \\
Water Index        & 0.33                                                                & \textbf{0.40}                                                                  & KNN                                                                       & R² (↑)          \\ \bottomrule
\end{tabular}
}
\vspace{4mm}
\caption{\textbf{Preliminary Results}. We show initial results on a range of tasks with various data modalities. The results in the left-hand column are achieved with a single model architecture, while the results on the right-hand column are achieved with domain-specific architectures from SustainBench.  We see our approach achieves competitive performance across nearly all of the tasks.}
\label{tab:results}
\end{table}

\vspace{-0.8cm}
\section{Discussion and Outlook}

As presented here, our method already satisfies the desiderata in Step 1: the model can process different sensor input modalities and resolutions, shares the same backbone architecture and can be equipped with different task heads.  Importantly, this model forms the starting point for experimentation with (i) multi-task learning which implies training a shared backbone on multi-modal data and various task-heads (Step 2); (ii) pre-training such a model backbone via any suitable self-supervised learning objective (Step 3); and (iii), a multi-modal learning component via dedicated tokenizers that can infuse arbitrary further geospatial databases (Step 4 and Step 5).

There are several challenges moving from the model we present towards a multi-task, multi-modal geospatial model that is capable of strong zero-shot performance on a broad range of tasks and modalities. First, the remote-sensing domain is fundamentally different to natural images, in that information content can be sparse but concentrated. Pragmatically, learning a common spatiotemporal index on this data may require new architectural innovation. Second, learning a model that generalizes across time can be difficult and require the model to extrapolate. Further, it will require us to effectively model distributions over possible futures. Finally, a non-trivial amount of resources will be required to acquire the dataset and compute resources necessary to train a model to interpolate spatially.

\bibliographystyle{plainnat}
\bibliography{bib}

\section{Appendix}

\subsection{Architectural Details}
\subsubsection{Tokenizer}

\begin{figure}[htb]
    \centering
    \includegraphics[width=\textwidth]{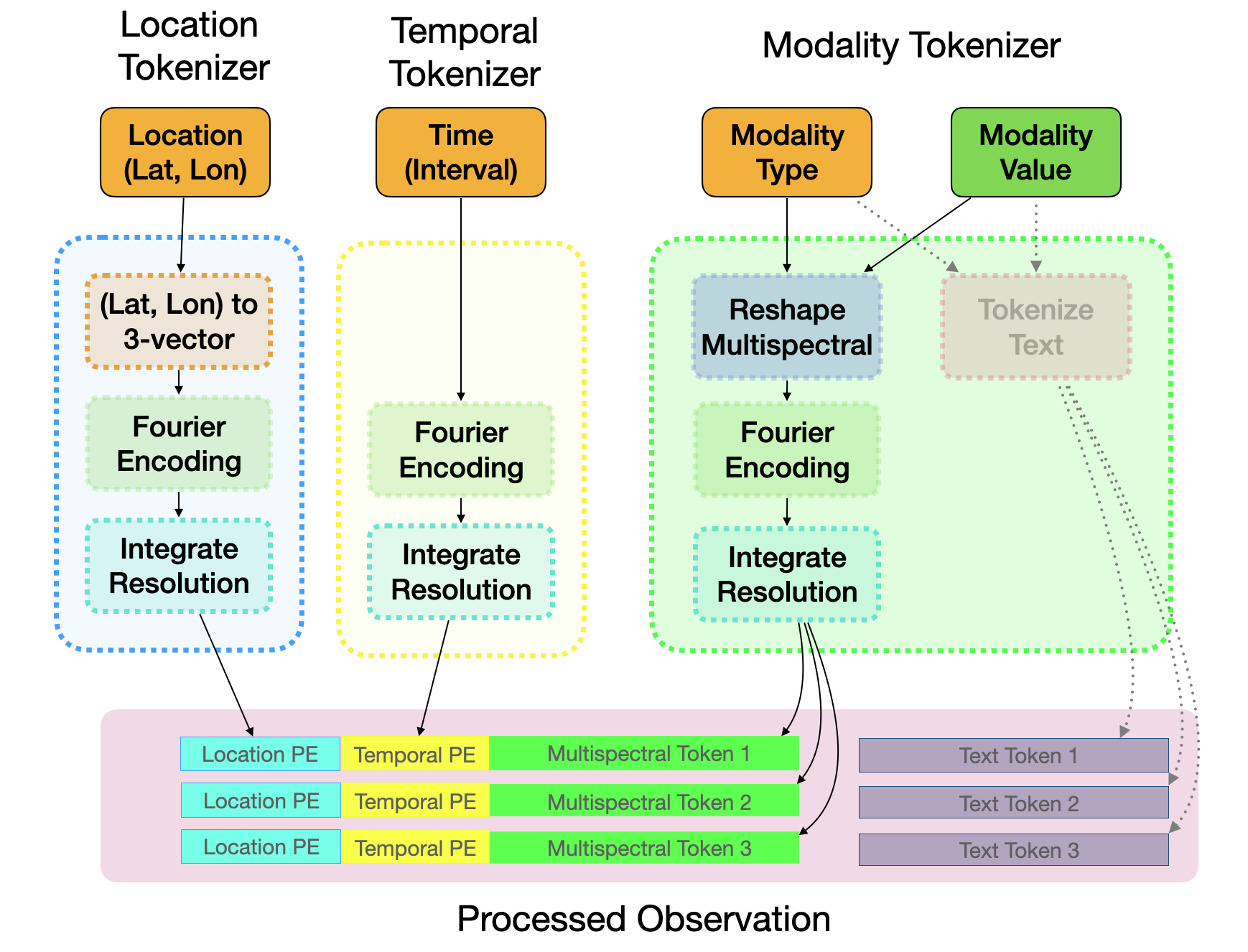}
    \caption{\textbf{Multispectral Modality Encoding}. The processed observation of a multispectral observation is a sequence (in raster order) of tokens. Each token is a concatenation of a Fourier encoded location and time with a normalized multi-spectral patch. In this example, the modality type would be a multispectral object type defining the number of spectral channels and their radiometric resolution as well as the image dimension. This type information determines which modality tokenizer to apply, and parameterizes its Fourier positional encoding. We grey out the text tokenizer given that this observation does not contain text.  }
    \label{fig:gamma-modality}
\end{figure}
As mentioned in the main text, the tokenizer serves the purpose of homogenizing diverse data-modalities by transforming them into a set of vectors. There are, however, several important aspects that warrant careful consideration.  Given the diversity of geo-spatial data, a key design decision is how to encode it as input for the downstream backbone, i.e. transform it to a set of vectors. For example, how should we encode a multi-spectral satellite image or time-series with its associated location and the corresponding uncertainty? The approach we take comprises three steps, under the assumption that the provided input can be packed in to a tensor of shape $T \times H \times W \times L$, where $H \times W$ is the spatial extent of the image, $L$ is the number of spectral bands, and $T$ is the number of frames in the time series ($T = 1$ for static images). 
We downsample the image or frame corresponding to each band with a stack of three convolutional layers. These layers do not mix information between bands or timesteps, and different datasets may share these layers if they share spectral bands. The result is a set of stacks of learned (spatial) features packed in to an array of shape $T \times H' \times W' \times C \times L$, where $H' \times W'$ is the spatial extent of the feature map, $C$ is the dimension of learned features. We add spatial positional encodings along the $H \times W$ dimensions, temporal positioanl encoding along the $T$ dimension, and spectral positional encodings along the $L$ dimension. Finally, we flatten this array to a matrix of shape $(T \cdot H' \cdot W' \cdot L) \times C$, where $T \cdot H' \cdot W' \cdot L$ corresponds to the number of tokens, each of dimension $C$. The overall tokenizer is visualized in Figure \ref{fig:gamma-modality}.

\textbf{Fourier Encoding.} The positional encodings use a modified version of the Fourier encoding, as used in \citep{perceiverio} and earlier. The traditional approach entails mapping a scalar $x$ to $\mathbb{R}^n$ as: 
\begin{equation}
    \gamma(x) = \left[
    \sin \left(2\pi k_1 x \right), \sin \left( 2\pi k_2 x \right), ... \sin \left( 2 \pi k_{\nicefrac{n}{2}} x \right), \cos \left( 2\pi k_1 x\right), ..., \cos\left( 2 \pi k_{\nicefrac{n}{2}} x\right) \right]
\end{equation}
where $k_1, ..., k_{\nicefrac{n}{2}}$ is a sequence of increasing frequencies and $n$ is even. The encoding of a vector in $\mathbb{R}^d$ is then simply a concatenation of the encodings of the scalar components, i.e.:
\begin{equation}
\label{eq:vec_fourier}
\mathbf{\gamma}(\mathbf{x}) = \left[\gamma(x_1), ..., \gamma(x_d)\right]
\end{equation}
We extend this scheme in two ways. First, we account for uncertainty $\Delta x$ in $x$ by considering the average of $\gamma$ in the interval $[x - \Delta x, x + \Delta x]$. More precisely: 
\begin{equation}
    \gamma(x \pm \Delta x) = \frac{1}{2\Delta x} \int_{x - \Delta x}^{x + \Delta x} \gamma(x') dx'
\end{equation}
We evaluate $\gamma(x \pm \Delta x)$ analytically with a symbolic math processing engine. Second, we discovered that the concatenation scheme of Equation~\ref{eq:vec_fourier} makes certain trade-offs that were sub-optimal for our use-case. Concretely, we found that functions defined on $\gamma(\mathbf{x})$ were less expressive along dimensions that were not axis-aligned. The reason is that Equation~\ref{eq:vec_fourier} implicitly assumes that the frequency vectors are axis aligned, i.e. they lie either on the x or the y axis of the Fourier plane. To account for this, we modified the multi-dimensional Fourier encoding scheme to be: 
\begin{equation}
    \gamma(\mathbf{x}) = \left[
    \sin \left(2\pi \mathbf{k_1} \cdot \mathbf{x} \right), \sin \left( 2\pi \mathbf{k_2} \cdot \mathbf{x} \right), ... \sin \left( 2 \pi \mathbf{k_{\nicefrac{n}{2}}} \cdot \mathbf{x} \right), \cos \left( 2\pi \mathbf{k_1} \cdot \mathbf{x}\right), ..., \cos\left( 2 \pi \mathbf{k_{\nicefrac{n}{2}}} \cdot \mathbf{x}\right) \right]
\end{equation}
Here, $\mathbf{k}_{(\cdot)} \cdot \mathbf{x}$ denotes the inner product, and $\mathbf{k}_1, ..., \mathbf{k_{\nicefrac{n}{2}}}$ is a sequence of vectors lying in the first quadrant of the Fourier plane. 

\subsubsection{Task Heads}
A task head consumes the latent representations produced by the model and produces an output. We consider three types of task heads in this work, for three different tasks: classification, regression and semantic segmentation. 

For classification and regression, the computation proceeds as follows. The set of latent representation vectors (as produced by the backbone) are used to produce sets of keys and values for a cross-attention mechanism, whereas the query is a single learned parameter vector. The output of the cross attention mechanism is then a single vector, which is then passed through a MLP to produce either the logits (for classification) or the regressed predictions (for regression). 

For segmentation, the same set of latent representation vectors (from the backbone) are used to produce sets of keys and values. The queries are produced by Fourier encoding the position of a pixel of interest in the output segmentation image. This yields $H \cdot W$ queries (where the segmentation map is a $H \times W$ image), and equally as many output vectors from the cross attention mechanism. Each vector is processed by a 2 layer MLP (in parallel), and the output is reshaped to a $H \times W$ image.

\subsection{Challenges around Batching of Samples}
Setting up the model to efficiently support diverse modalities brings its own unique challenges. One salient challenge has to do with batching of different data modalities when training on multiple diverse datasets simultaneously (as required for Step 2). Different data samples (each originating from different and diverse datasets) may result in a different number of tokens. However, in order to utilize GPUs effectively, these token arrays must be stacked together in a batch before being passed in to the backbone. 

The typical strategy to address this is to use padding in conjunction with attention masking. For example, if the sample at index 0 has 100 tokens whereas that at index 1 has 150 tokens, the former will be padded by 50 tokens such that the overall batch has shape $2 \times 150 \times C$ (where $C$ is the number of feature dimensions). The padding tokens are subsequently masked away while computing the attention scores, and they do not contribute to the overall result. 

However, this presents a challenge that arises while training multiple modalities: if the sample at index 1 would instead have $4000$ tokens, sample 0 must be padded by 3900 tokens that do not contribute to the computation but still require memory and FLOPs. This problem exacerbates for large batch sizes containing diverse samples, where it is likely that a single sample with a large number of tokens (up to 15k tokens, in our case) necessitates excessive padding (and therefore wasted compute) for most other samples. To address this issue, we devise the following strategy. Before passing the tokens to the backbone, we cluster samples by their respective number of tokens. Within each cluster, all samples have a similar (but not the same) number of tokens. We process all tokens within a single cluster in a batched way, but the different clusters are processed in different batches. We use DBSCAN to carry out the clustering, which allows us to specify the maximum number of padding tokens that we wish to allow within a cluster, but the number of clusters is deduced automatically. The choice of the number of padding tokens per cluster determines the trade-off between memory requirement and speed, since a single cluster will be fastest due to efficient parallelization on the hardware level. 

\end{document}